\documentclass[10pt,twocolumn,letterpaper]{article}

\usepackage[pagenumbers]{iccv}     

\definecolor{iccvblue}{rgb}{0.21,0.49,0.74}
\usepackage[pagebackref,breaklinks,colorlinks,allcolors=iccvblue]{hyperref}

\def\paperID{15015} 
\def\confName{ICCV}
\def\confYear{2025}

\title{Words as Geometric Features: Estimating Homography using \\ Optical Character Recognition as Compressed Image Representation}

\author{\small{Ross Greer}\thanks{All authors are with the University of California, San Diego. Ross Greer is with the University of California, Merced.}\\
{}
\and
\small{Alisha Ukani}\\
{}
\and
\small{Katherine Izhikevich}\\
{}
\and
\small{Earlence Fernandes}\\
{}
\and
\small{Stefan Savage}\\
{}
\and
\small{Alex C. Snoeren}\\
{}
}

\begin{document}
\maketitle
\begin{abstract}
Document alignment and registration play a crucial role in numerous real-world applications, such as automated form processing, anomaly detection, and workflow automation. Traditional methods for document alignment rely on image-based features like keypoints, edges, and textures to estimate geometric transformations, such as homographies. However, these approaches often require access to the original document images, which may not always be available due to privacy, storage, or transmission constraints. This paper introduces a novel approach that leverages Optical Character Recognition (OCR) outputs as features for homography estimation. By utilizing the spatial positions and textual content of OCR-detected words, our method enables document alignment without relying on pixel-level image data. This technique is particularly valuable in scenarios where only OCR outputs are accessible. Furthermore, the method is robust to OCR noise, incorporating RANSAC to handle outliers and inaccuracies in the OCR data. On a set of test documents, we demonstrate that our OCR-based approach even performs more accurately than traditional image-based methods, offering a more efficient and scalable solution for document registration tasks. The proposed method facilitates applications in document processing, all while reducing reliance on high-dimensional image data.
\end{abstract}    
\section{Introduction}
\label{sec:intro}

Document alignment and registration are critical tasks in many real-world applications, such as automated form processing \cite{casey1992intelligent, baviskar2021efficient}, anomaly detection \cite{diarra2024doc}, and workflow automation \cite{baviskar2021efficient, fischer2021multi}. Traditional methods for document alignment rely on image-based features, such as keypoints, edges, or textures, to compute geometric transformations like homographies \cite{doermann2003progress, kornfield2004text, tang2023unifying, martin2023sealclub}. However, these approaches often require access to the original document images, which may not always be available due to storage constraints, privacy concerns, or transmission limitations \cite{aura2006scanning, small2012storing, ion2011home}. 

\begin{figure}
    \centering
    \includegraphics[width=0.94\linewidth]{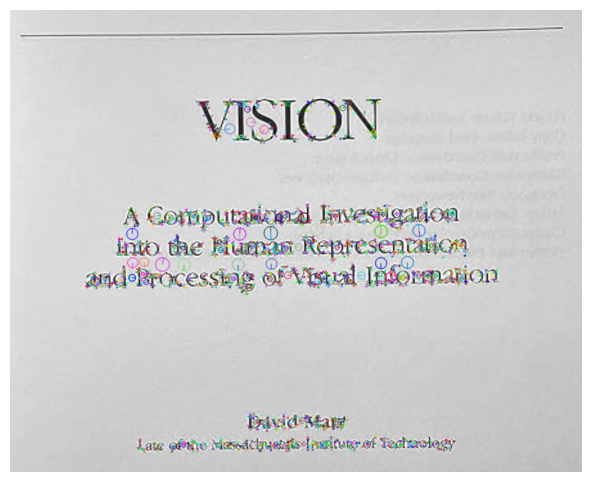}
    \includegraphics[width=0.94\linewidth]{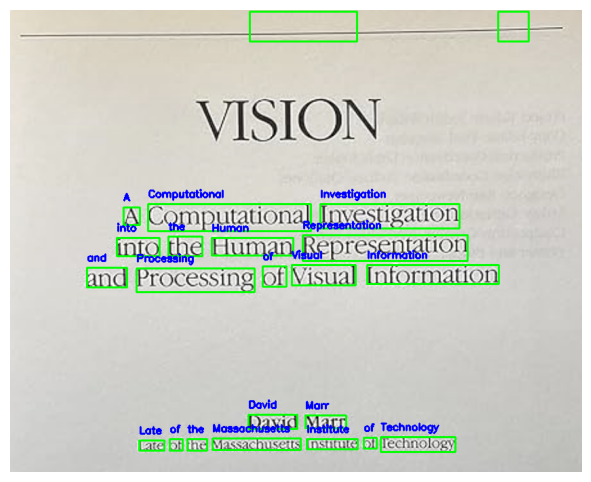}
    \caption{SIFT features (top) and OCR features (bottom) on a document. Stated as a variant of Marr's definition of vision, here we introduce the idea that ``knowing what \textbf{\textit{word}} is where by looking", as captured by current and imperfect OCR models, is sufficient for the estimation of homography matrices relating the geometric transforms between pairs of documents for purposes of alignment for downstream vision tasks.}
    \label{fig:intro}
\end{figure}

This paper proposes a novel approach to document alignment by leveraging Optical Character Recognition (OCR) outputs as features for computing homographies. Instead of relying on pixel-level image data, our method uses the spatial positions and textual content of recognized words to estimate the transformation between documents. This approach is particularly advantageous in scenarios where only OCR outputs are preserved, as it eliminates the need for storing or transmitting full document images. Furthermore, our method is robust to OCR noise, as it can employ robust estimation techniques (in our implementation, RANSAC [Random Sample Consensus] \cite{fischler1981random}) to handle outliers and inaccuracies in the OCR output. The proposed method has broad applicability, and by aligning documents based on OCR-derived features, we enable efficient and scalable processing of textual data without compromising on accuracy or robustness.

\section{Related Research}

\subsection{Homography and Document Tasks in Computer Vision}

Document alignment has been extensively studied in computer vision and document analysis \cite{tang1999document, tang2005document, lecun2007energy}. For the purposes of this research, we assume a photographed, scanned, or digital document which has a dual template that may or may not have alterations (e.g., a completed and signed form, or a document with varying field values). We further note that the two documents need not be simultaneously available. On the assumption that both documents are planar, traditional methods of aligning such documents involve estimating the homography matrix \( \mathbf{H} \), which represents a projective transformation that relates the coordinates of points between two images of the same planar surface \cite{hartley2003multiple}. It is a \( 3 \times 3 \) matrix defined as:
\[
\mathbf{p'} = \mathbf{H} \mathbf{p}
\]
where \( \mathbf{p} = [x, y, 1]^T \) and \( \mathbf{p'} = [x', y', 1]^T \) are the homogeneous coordinates of a point in the original and transformed image, respectively. The homography matrix \( \mathbf{H} \) is parameterized by 8 independent parameters, as it has 9 elements in total, but the scale factor is typically normalized. These parameters can be represented as:
\[
\mathbf{H} = \begin{bmatrix}
h_1 & h_2 & h_3 \\
h_4 & h_5 & h_6 \\
h_7 & h_8 & 1
\end{bmatrix}
\]
where \( h_1, h_2, \dots, h_8 \) are the parameters that define the transformation between the two images. Estimation methods often rely on image-based features, such as SIFT (Scale-Invariant Feature Transform) \cite{lowe1999object}, SURF (Speeded-Up Robust Features) \cite{bay2006surf}, or ORB (Oriented FAST \cite{rosten2006machine} and Rotated BRIEF \cite{calonder2010brief}) \cite{rublee2011orb} to detect and match keypoints between images, applied often in both document processing \cite{martin2023sealclub} and general computer vision. These methods have been successfully applied to tasks like image stitching \cite{brown2007automatic}, document registration \cite{safari1997form}, and form processing \cite{casey1992intelligent}. However, they require access to the original image data, which can be a limitation in many practical scenarios when data privacy, integrity, or storage may be a concern.

\begin{figure*}
    \centering
\includegraphics[width=.27\textwidth]{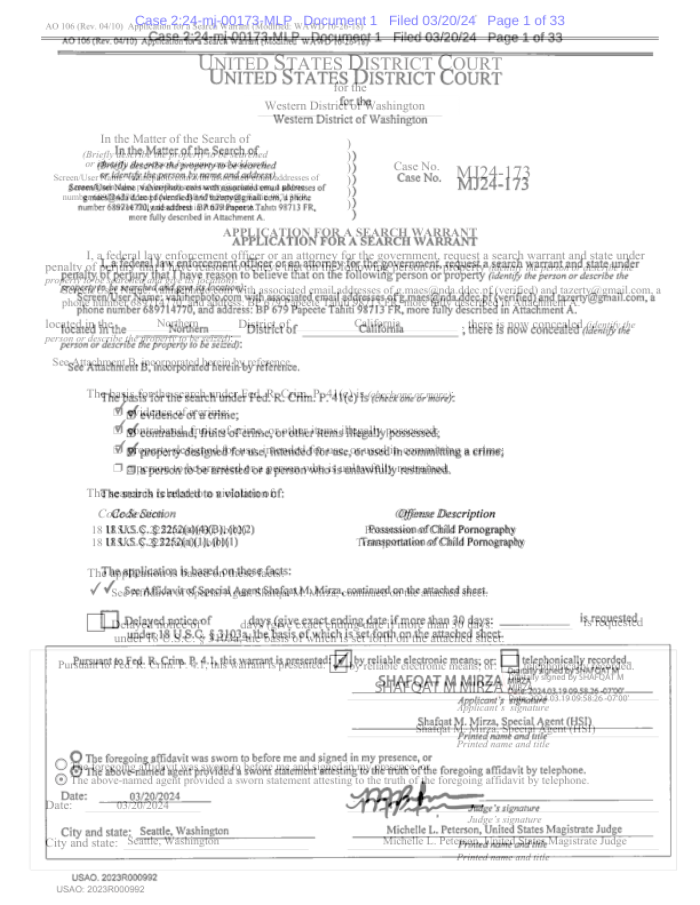}
\includegraphics[width=0.26\textwidth]{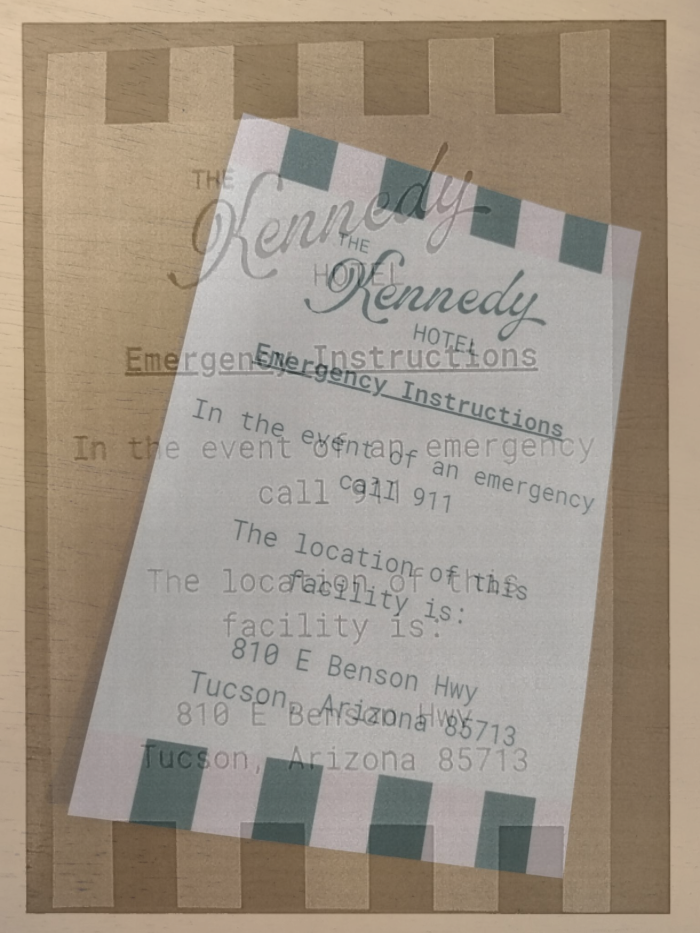}
\includegraphics[width=0.45\textwidth]{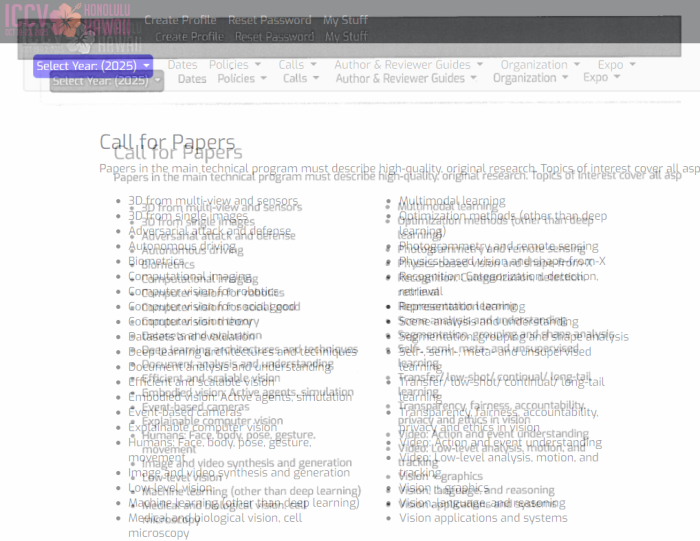}
    
    \caption{Example of documents before image alignment through homography estimation. Corresponding fields are in different locations, confounding document analysis. These samples correspond to the court document (digital and scanned), flyer (standard scanned and perspective), and web printout (digital and scanned) described in the Experimental Evaluation section.}
    \label{fig:prealign}
\end{figure*}

Though no OCR system is perfect, many large, robust OCR models are available under varying production levels, such as the free and open-source Tesseract software\footnote{https://github.com/tesseract-ocr/tesseract}, and proprietary models such as Google Document AI\footnote{https://cloud.google.com/document-ai}, Amazon Textract\footnote{https://aws.amazon.com/textract}, and Adobe OCR\footnote{ https://developer.adobe.com/document-services/docs/overview/}. Advances in OCR technology have provided high accuracy in text recognition and layout analysis; related research has explored the use of OCR outputs for tasks like document text alignment and information extraction \cite{de2023evaluating, mitra2000information, yalniz2011fast}. However, the use of OCR outputs as features for geometric transformation estimation remains underexplored, likely due to the general success of gradient-based features (e.g. SIFT). In prior research on document-based homography estimation, OCR has been integrated in correspondence matching algorithms in prior literature, but only as a mechanism for reducing the search region \cite{fan2016automatic} or identifying distinct structural keypoints on imaged characters \cite{mahajan2019character}. In these cases, the matches are still made on gradient-based visual features as in traditional correspondence matching, rather than making use of the special constraints afforded by the nature of text document data. To our knowledge, we are the first to use the natural language human-readable glyphs themselves as the set for encoding the image correspondence features. A benefit of doing so is that the problem resolves from a continuous open-set of descriptors to a discrete set of descriptors, simplifying the matching procedure.

\subsection{Comparison of OCR to SIFT features}
In certain document registration applications, it is possible that pairwise sets of images which must be aligned may not be available simultaneously. In these cases, it is not possible to compute the homography matrix, which would be the most efficient representation of the geometric relationship between the documents with only 9 parameters. An alternative is to compress and store features such as SIFT keypoints and descriptors, which are known to be particularly effective in correspondence matching. A typical SIFT descriptor is 128-dimensional per keypoint, and an image can have hundreds to thousands of keypoints. As a back-of-envelope calculation, if an image has 500 keypoints, these features would be represented by 64,000 floats, and at 4 bytes per float, we require around 256 KB per image, with more keypoints leading to an even larger file size. Besides the descriptors, SIFT also stores the location of each keypoint, typically as 2D coordinates (x, y) in the image. For each keypoint, this requires 2 additional floating-point numbers (4 bytes per float), raising the total SIFT information to 130 floats per feature, or $\sim 260$ KB for our hypothetical image. 

By comparison, we now repeat the estimate with the output of an OCR algorithm taken over a document image. If an image has 200 words, and each word is 5 characters on average, we have around 1,000 characters. Assuming ASCII characters stored as UTF-8, at an average of one byte per character, we can represent this with $\sim 1 $ KB. Combining this with four float parameters describing each of the 200 words, we reach 800 floats at 4 bytes each, giving $\sim 3.2$ KB, for a total OCR size of $ \sim 4.2$ KB, for a compression ratio of approximately 50. In this research, we propose that due to the inherent visual structure of written language, the information preserved in the OCR is adequate to estimate the homography, rather than relying on the domain generalizable but higher dimensional features of gradient-based descriptors like SIFT.

\section{Methodology}

The core of our methodology is replacement of traditional correspondence keypoints and feature descriptors in the homography estimation algorithm, and computation of feature similarity as a string-matching task. Specifically,
\begin{enumerate}
    \item gradient-based scale-space extrema, or similar keypoint candidates, are replaced by the centroids of OCR-detected words, and 
    \item high-dimensional gradient-information descriptors are replaced by the OCR-extracted textstring.
\end{enumerate}

This feature extraction is our primary step, which compresses the document and converts its representation from image space to text space. Further, we show that while this compression discards image information, it preserves the information relevant to map homography to a later-available template image, facilitating tasks in document alignment and understanding. The challenge of non-alignment is illustrated in Figure \ref{fig:prealign}. 

\begin{figure*}
    \centering
\includegraphics[width=.4\textwidth]{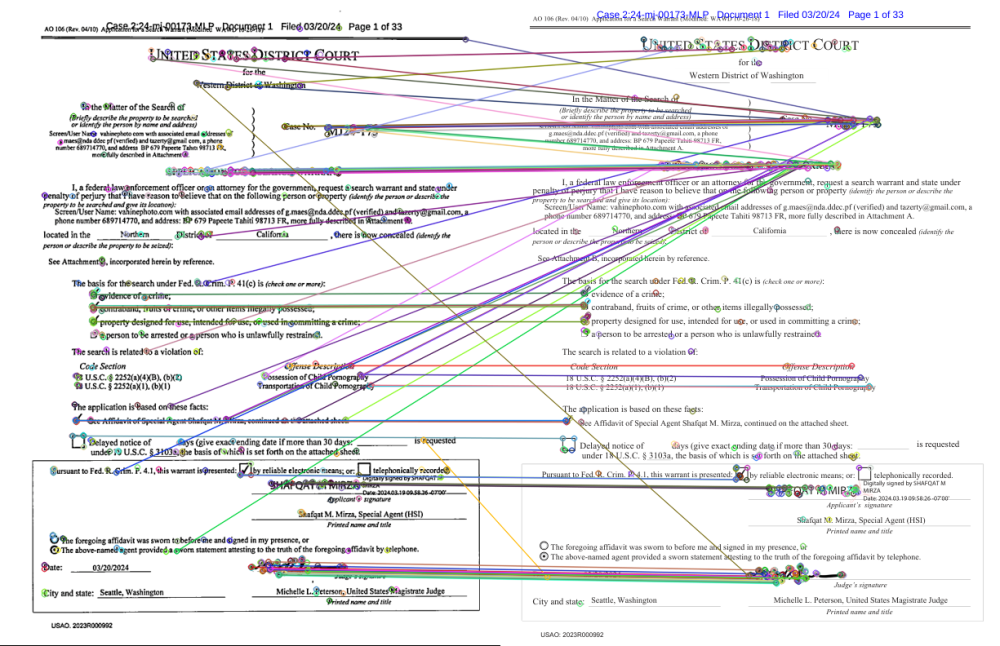}
\includegraphics[width=.59\textwidth]{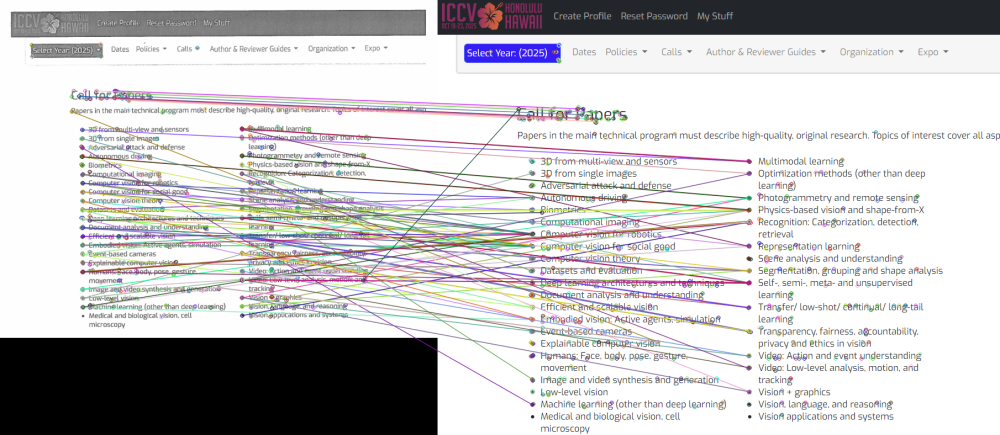}
\includegraphics[width=.4\textwidth]{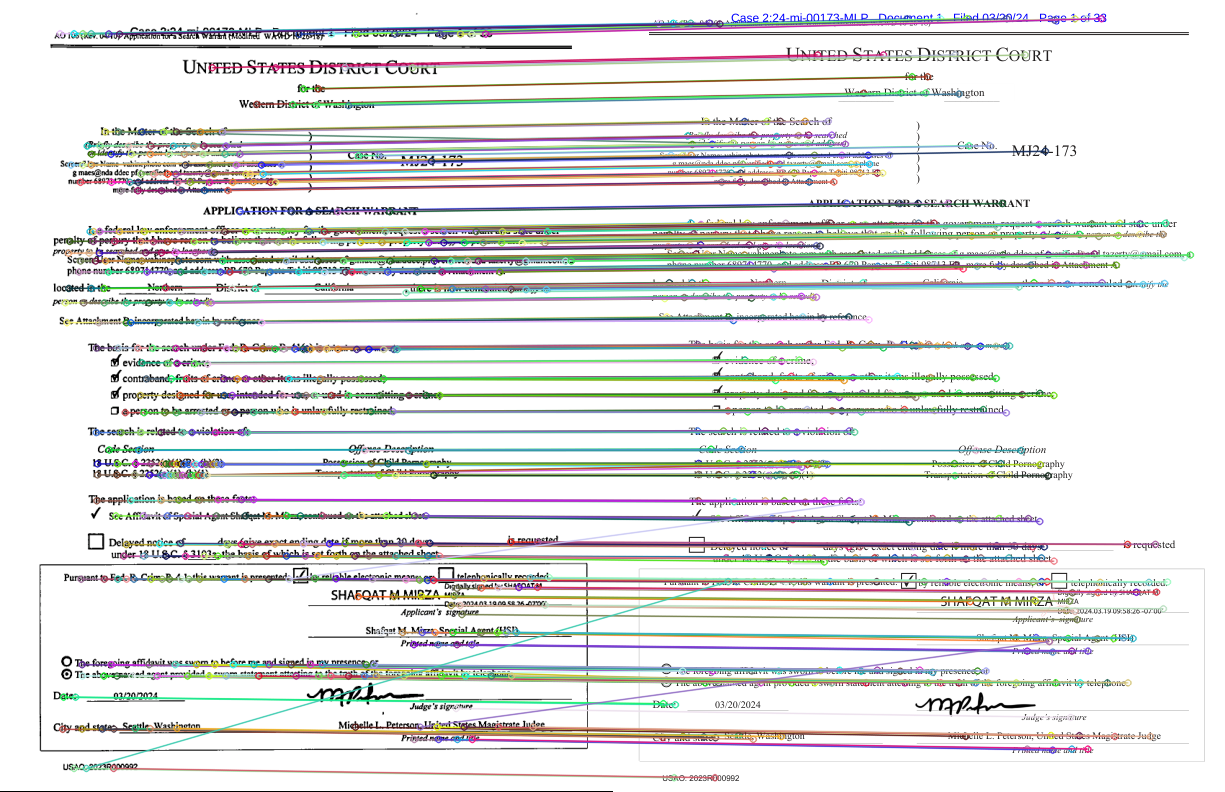}
\includegraphics[width=.59\textwidth]{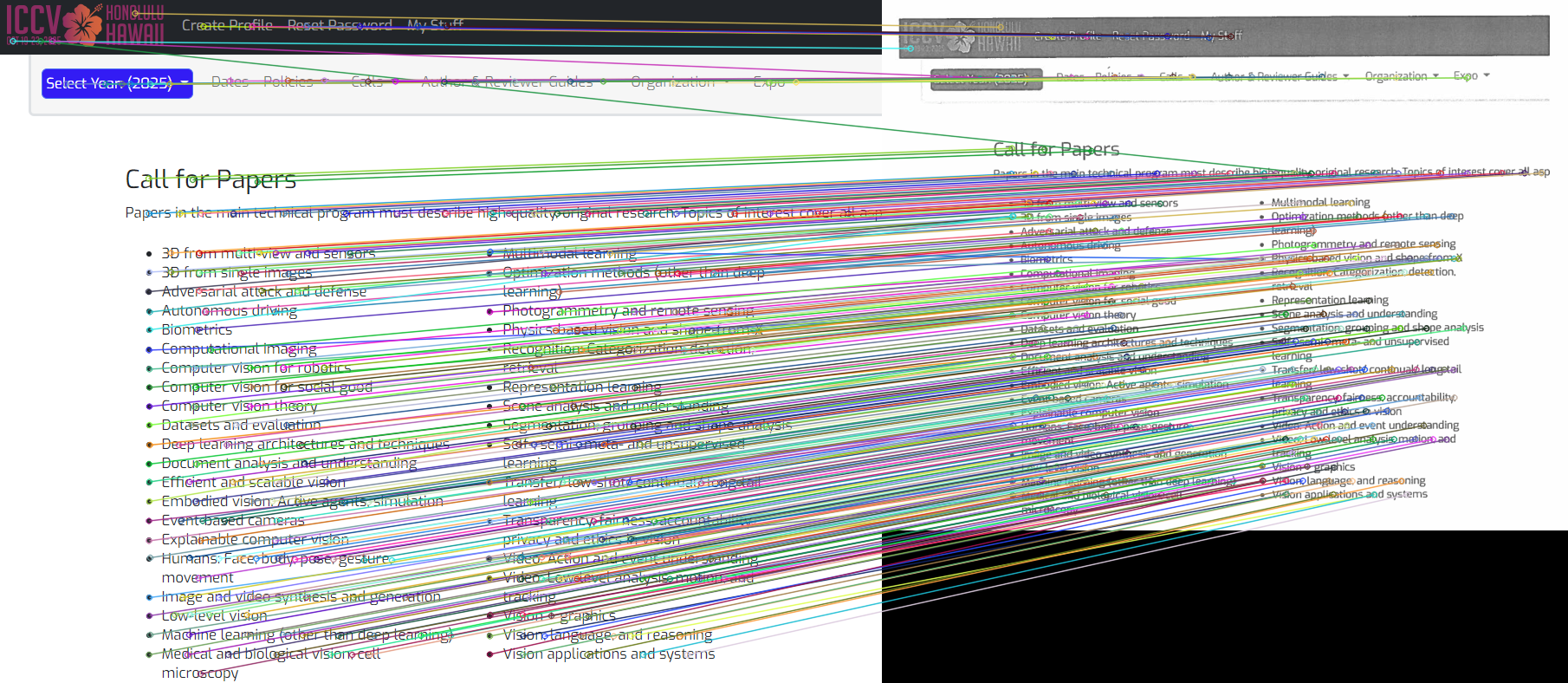}
    \caption{SIFT matches (top) vs. OCR matches (bottom) for two sample pairs of documents. There are many spurious SIFT matches, though this is not too problematic given the robustness of RANSAC estimation techniques. However, the strengths of OCR-driven features are clear by comparison.}
    \label{fig:matches}
\end{figure*}

\begin{figure}
    \centering
    \includegraphics[width=.45\textwidth]{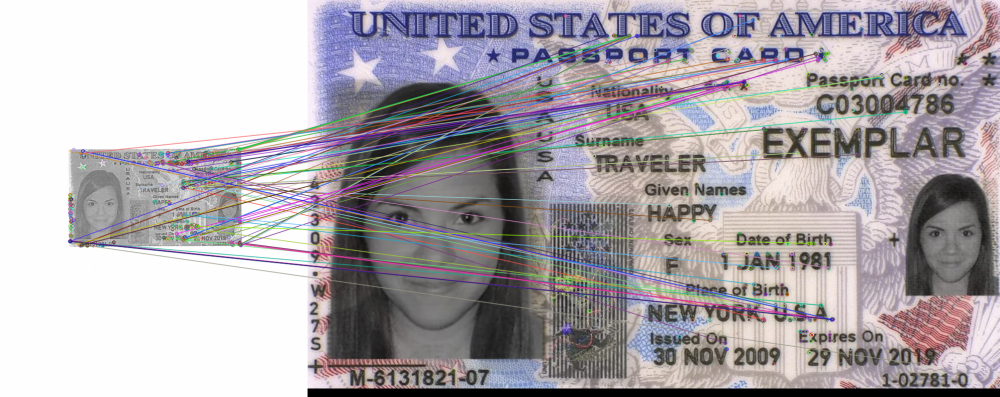}
    \includegraphics[width=.45\textwidth]{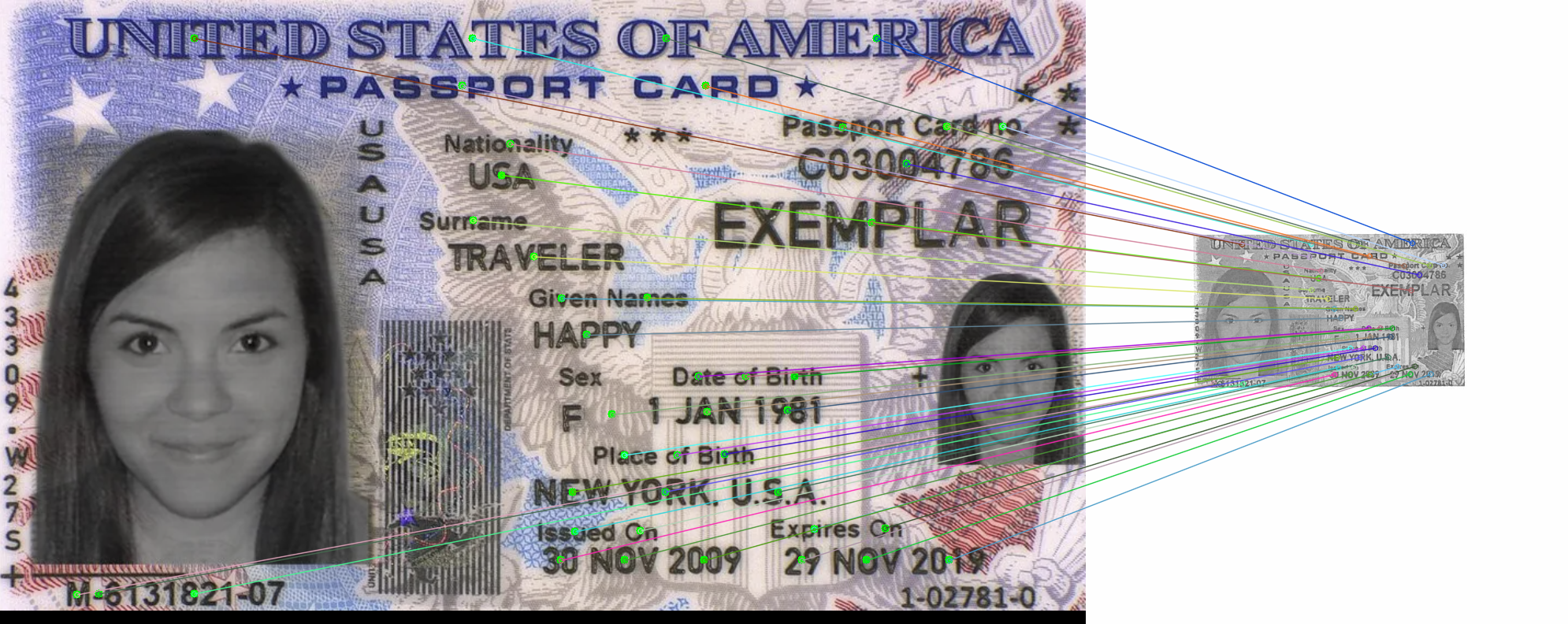}
    \caption{SIFT matches (top) vs. OCR matches (bottom) for the passport card document described in the Experimental Evaluation section. The SIFT method creates many incorrect crossings, which eventually lead to an incorrect estimate of \textbf{H}.}
    \label{fig:matchespassport}
\end{figure}

\begin{figure}
    \centering
    \includegraphics[width=.29\textwidth]{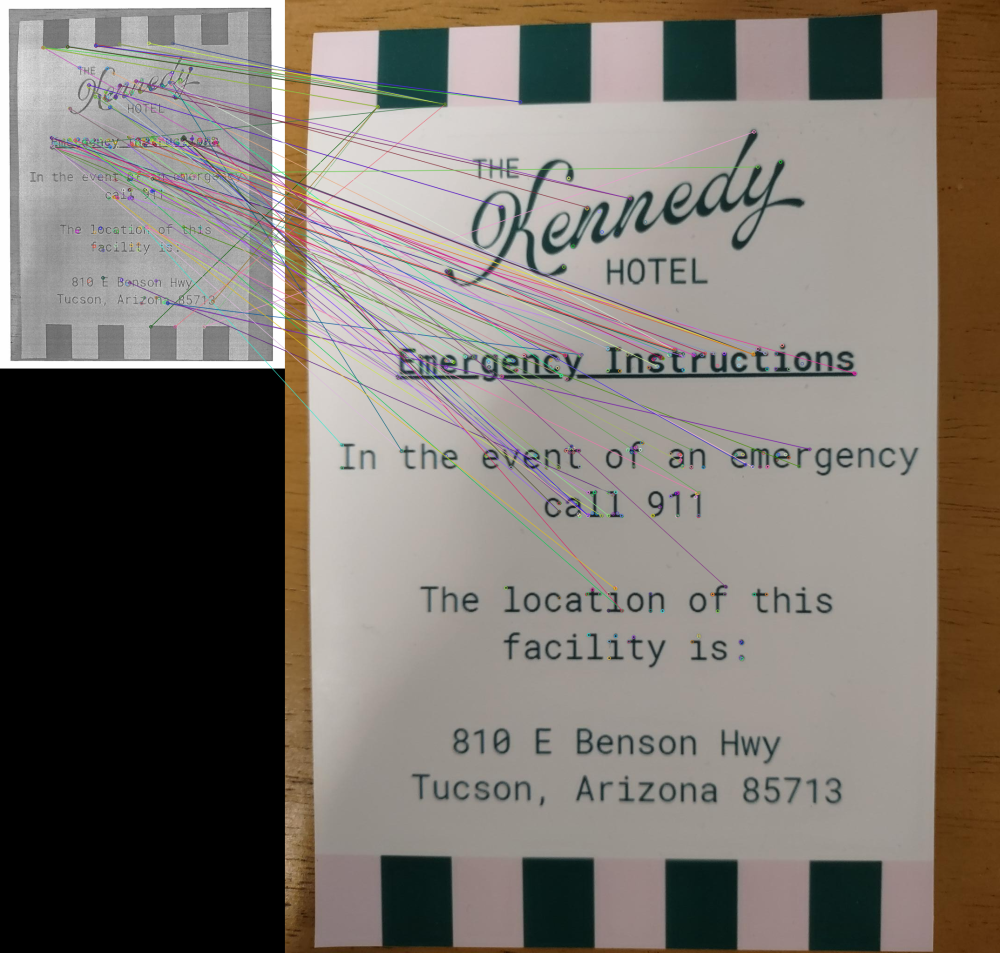}
    \includegraphics[width=.29\textwidth]{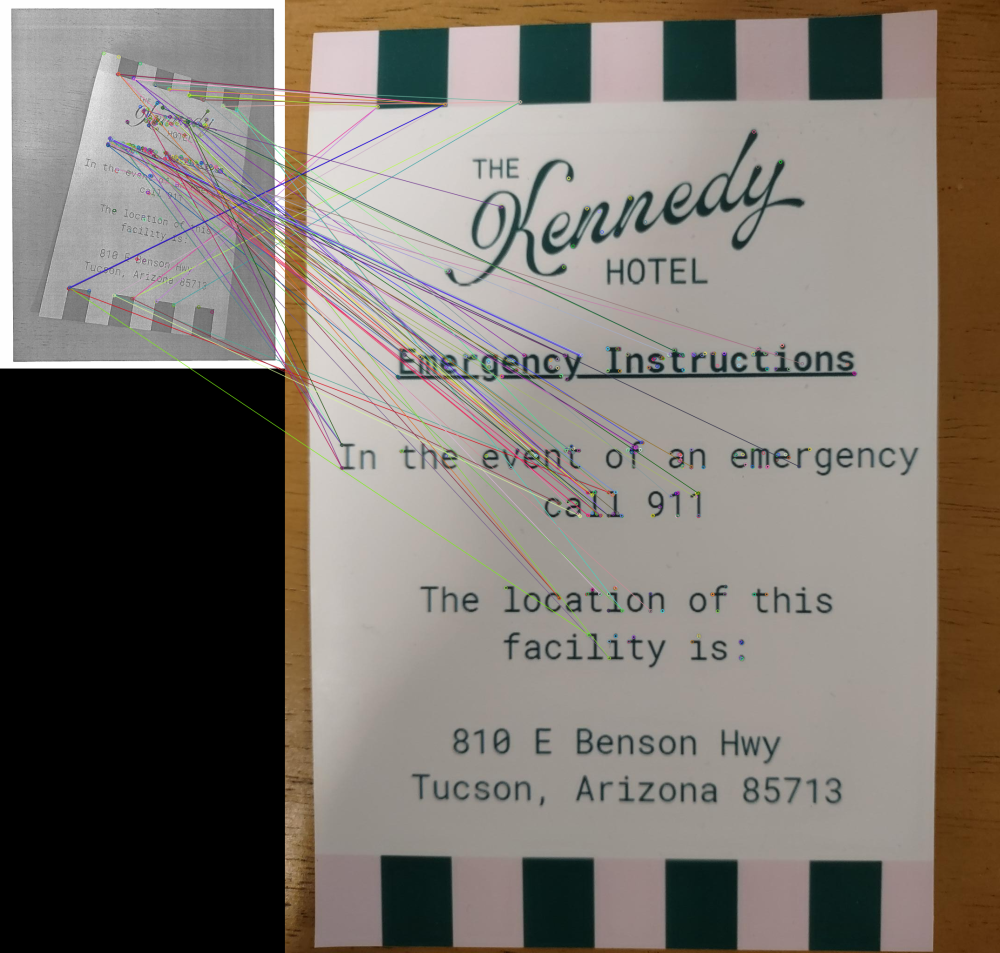}
\includegraphics[width=.29\textwidth]{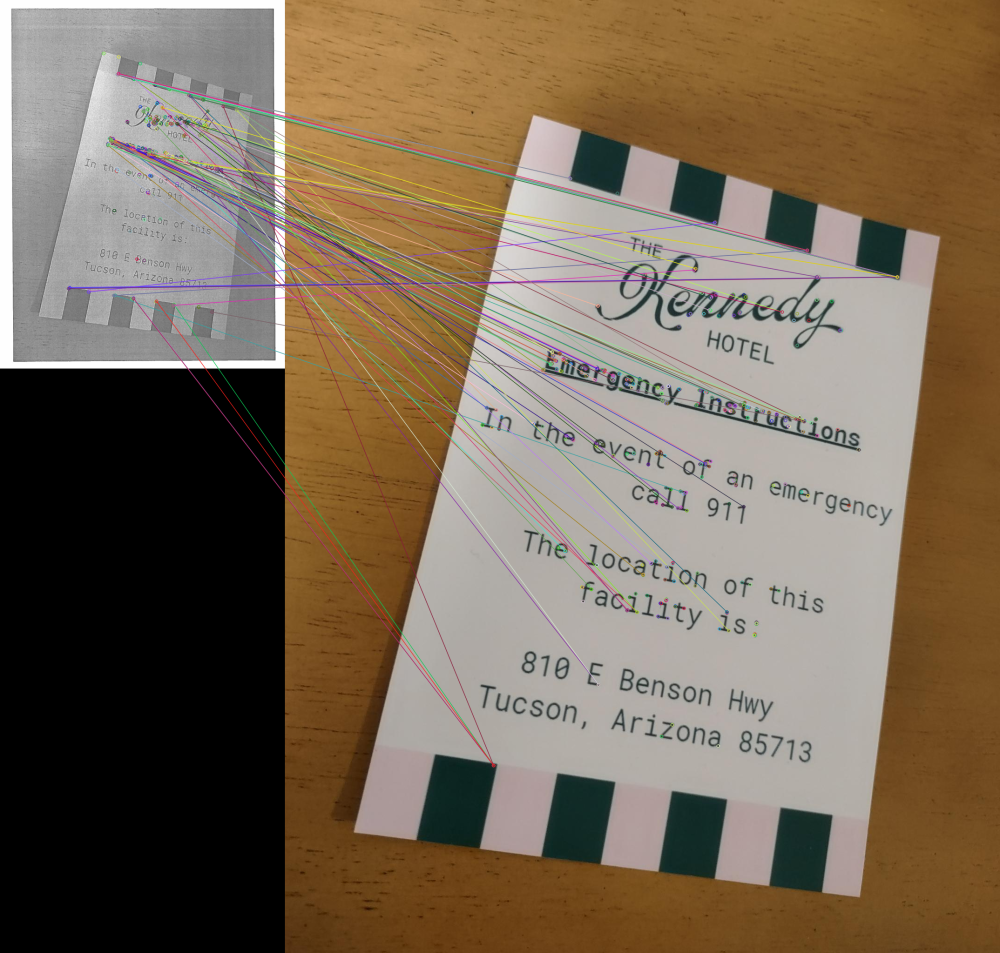}
    \includegraphics[width=.29\textwidth]{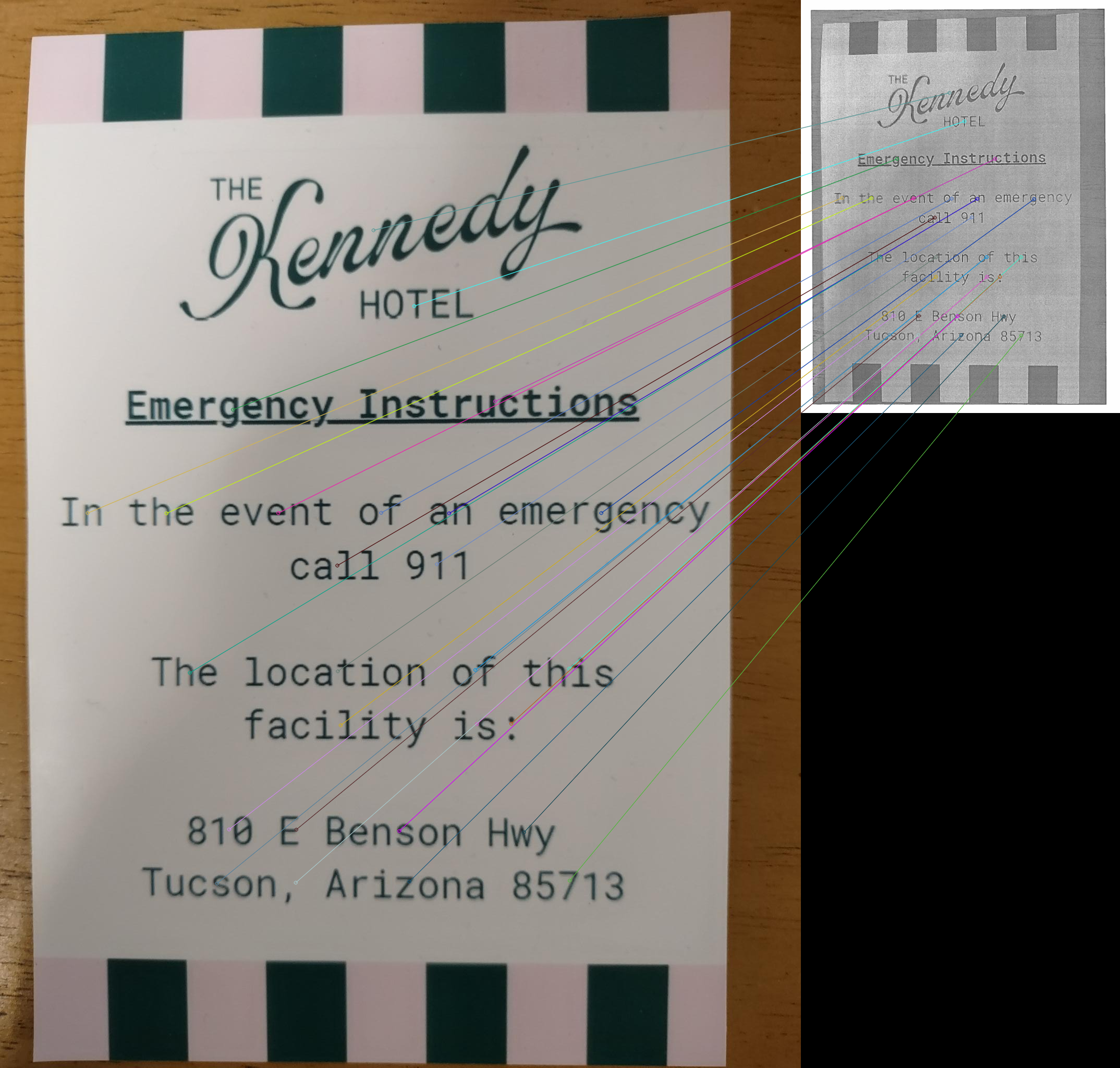}

    \caption{SIFT matches (top three images) vs. OCR matches (bottom) for sample pairs of the `flyer' document. Gray images are 200dpi scans, and for our experiments we evaluate at both a direct and indirect perspective in multiple combinations. We note the presence of confounding matches in the SIFT features, compared to a clear relationship on the OCR features despite the relatively limited number of words available.}
    \label{fig:matches_flyer}
\end{figure}

Once such a template document becomes available, our second step involves feature matching between the OCR output of both documents. After filtering for stopwords and whitespace, which may have many co-occurences in close proximity, we perform a bipartite string matching between words found in each document to establish correspondence. Feature matching results comparing SIFT and OCR features are illustrated in Figures \ref{fig:matches}-\ref{fig:matches_flyer}. When multiple instances of a candidate word are found in the template, we use a heuristic selection of the minimal pixel distance, under the assumption that most transforms will not invert this ordering. On the other hand, we suggest that it is still reasonable to drop this assumption and either omit such words or even allow them to pair with the ``incorrect" matches, as the use of RANSAC will, with tuned confidence, algorithmically determine the correct homography regardless of correspondence noise. Our selected data in the Experimental Evaluation section contains both types of instances, where this assumption does and does not hold, demonstrating the robustness of RANSAC to such correspondence noise.

Following this OCR processing, feature extraction, and matching, the traditional homography estimation pipeline continues. Using the matched word pairs, we compute a homography matrix that maps one document's coordinate system to the other. Since OCR outputs may contain noise and outliers, we employ RANSAC to robustly estimate the homography. RANSAC iteratively selects random subsets of matches, computes the homography, and evaluates its consistency with the remaining matches. This process ensures that the final homography is robust to OCR measurement and string recognition errors and outliers. These methods are particularly relevant to our work, as OCR outputs can be inherently noisy due to recognition challenges, layout inconsistencies, or variations in document quality. 

As a note on the statistical confidence of this method and its tractability, consider the 99\% confidence table provided in Table \ref{tab:ransac_iterations}, built from the equation 
\[
N = \frac{\log(1 - p)}{\log(1 - (1 - e)^s)}, 
\]
where $N$ is number of iterations, $p$ is model confidence, $e$ is estimated error rate of matches, and $s$ is minimal number of samples required to estimate the model. The estimation of error rate is multifaceted, comprised not only of the OCR error rate, but also the precense of matching errors, many of which are addressed by removal of stopwords (which reduces the number of candidate matches). This may be relevant for documents which have only a few words, as at least four good matches are required for a good estimate. But, if it is the case that there are very, very few words on the page, unless there are printed images or other figures, it is likely that the page is mostly blank, which would provide similar challenges to gradient-based descriptors such as SIFT. One of our test documents, the flyer, contains very few words to test this condition.

\begin{table}
    \centering
    
    \begin{tabular}{ccc}
        \toprule
        Error Rate (\( e \)) & Required Iterations (\( N \)) \\
        \midrule
        10\%    & 9 \\
        30\%    & 40 \\
        50\%    & 150 \\
        70\%    &  2000 \\
        \bottomrule
    \end{tabular}
    \caption{Number of RANSAC iterations (\( N \)) needed to achieve 99\% confidence for different matching error rates for a 4-correspondence or 8 parameter model (e.g. \textbf{H}).}\label{tab:ransac_iterations}
\end{table}

By combining OCR-derived features with robust estimation, our method introduces a connection between the structure of text features and geometric alignment.

\begin{figure*}
    \centering
\includegraphics[width=.27\textwidth]{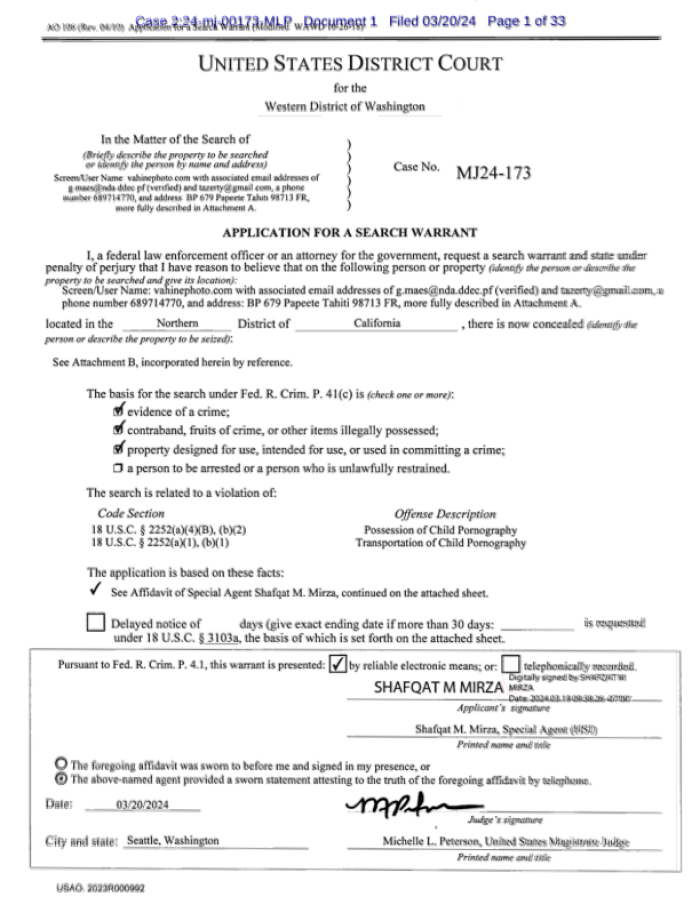}
\includegraphics[width=.26\textwidth]{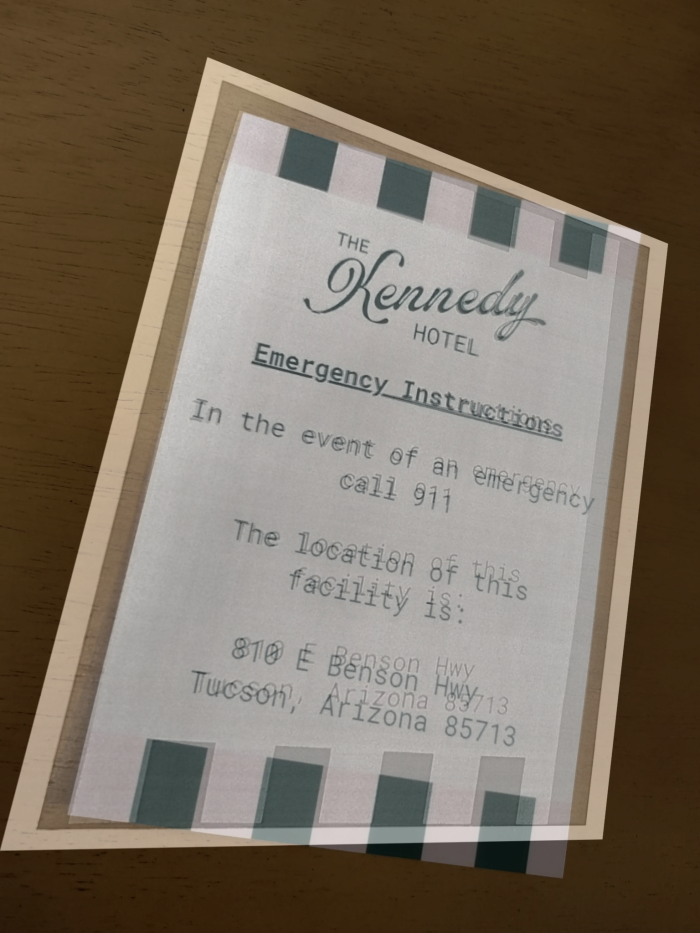}
\includegraphics[width=.45\textwidth]{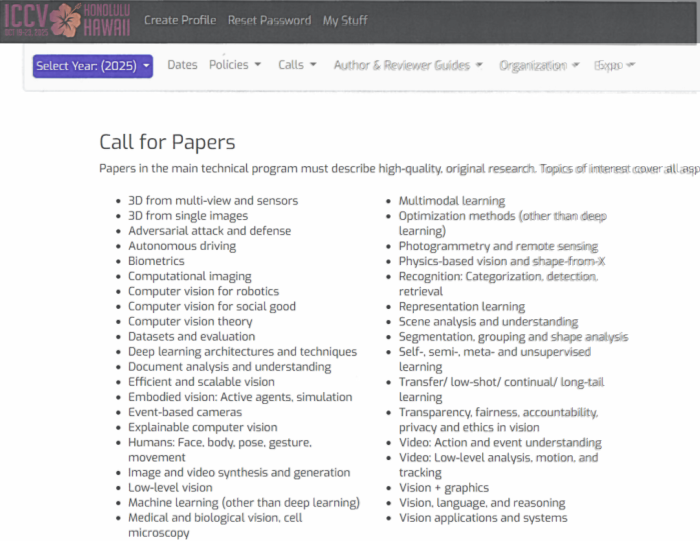}
    \caption{Documents aligned using OCR-feature method for estimating homography, with examples corresponding to those introduced in Figure \ref{fig:prealign}. In general, the OCR features are effective as a basis for performing these document alignments, even on challenging cases with limited word counts and strong changes of perspective.}
    \label{fig:result}
\end{figure*}

\section{Experimental Evaluation}


We evaluated our method on a dataset of digital and scanned documents, including printed media with varying inter- and intra-document fonts, as well as forms with handwritten or signed fields. The dataset was chosen to reflect a variety of document types and  layouts found in real-world use cases, with challenging quality levels, word sparsity, and distortions. All scans were taken at 200 DPI, the lowest quality available on our scanner, to assess performance under edge conditions. In particular, the dataset includes:
\begin{itemize}
    \item \textbf{Flyer and Perspective Flyer}: A small printed flyer with only 26 words, with an additional image taken at an angle to introduce a notable perspective transform.
    \item \textbf{Passport Card}: A wallet-sized sample passport card, with many icons and gradients in the background. We note that the scan process naturally introduces a high volume of padding to the passport card and flyer. 
    \item \textbf{Court Document}: A sample public court document with many words, checkboxes, horizontal and vertical lines, and form fields, which also includes signatures, stamps, and overlapping text.
    \item \textbf{Web Printout}: A call for papers printed from a conference website, which includes text on varying backgrounds and an icon. 
\end{itemize}

\begin{table*}[]

    \centering
    \begin{tabular}{c|c|c|c|c}
    Document Set & Document 1 Size & Document 2 Size & $\text{Error}_{SIFT}$ & $\text{Error}_{OCR}$ \\ \hline
         Court Document \& Scan & $2200 \times 1700$ & $2200 \times 1700$ & 1.5735 px  &  \textbf{1.2693 px} \\
    
     Passport Card \& Scan & $1378 \times 775$ & $2200 \times 1700$ & 10527.6336 px  &  \textbf{1.7814 px} \\
     Web Printout \& Scan & $1018 \times 788$ & $2200 \times 1700$ & 1.4389 px  &  \textbf{1.1925 px} \\
     Flyer \& Scan & $2048 \times 1536$ & $2200 \times 1700$ & \textbf{1.5654 px} & 2.4812 px \\
     Perspective Flyer \& Scan & $2048 \times 1536$  &$2200 \times 1700$ & 206.7182 px & \textbf{2.1958 px} \\ 
     Perspective Flyer \& Flyer Scan & $2048 \times 1536$  &$2200 \times 1700$ & \textbf{1.4364 px} & 2.0033 px\\
     Flyer \& Perspective Flyer Scan & $2048 \times 1536$ &$2200 \times 1700$ & 362.7059 px  & \textbf{2.9812 px}\\

    \end{tabular}
        \caption{Mean Symmetric Transfer Error (Square Root) for SIFT- and OCR-based Homography Estimation Approaches over a variety of test documents and their respective low-quality scanned versions. In some cases when there are very few words available on the document, the SIFT-based approaches may outperform the OCR-based approaches, but we note that such error rates are marginally similar.}
    \label{tab:error}
\end{table*}
We compare our OCR-based approach with traditional image-based methods using SIFT features. In addition to qualitative examples of success illustrated in the transformation of images from Figure \ref{fig:prealign} to Figure \ref{fig:result}, we also provide the symmetric transfer error of the estimated homography over the inlier correspondence set in Table \ref{tab:error}. 

Surprisingly, our OCR-as-features method often achieved even higher accuracy than the intensity gradient-based method. All homographies are notably low-error when using the OCR-based method, while the SIFT-based methods demonstrate failure on three of the trials. Despite the surprising result, this may have an intuitive explanation: SIFT features are most meaningful to applications in the embodied, natural, scenic world, rather than the sparse-signaled world of documents and printed words, where OCR has been tailored to excel at its specific recognition task, influencing the ability of these methods to describe and localize their respective images. This challenge for SIFT is most apparent in Figures \ref{fig:matches}-\ref{fig:matches_flyer}. 

Our OCR-driven method demonstrated strong robustness, successfully aligning documents even with OCR recognition and matching errors. Further, we again highlight that our method does not require the preservation of the original image beyond the extraction of initial OCR features.


We note a few limitations of this method. One limitation of the particular application examined in this paper, homography estimation, is the inability of the homography model to account for fold distortions \cite{martin2023sealclub}, since this violates the planar assumption. However, techniques such as local homography estimation can overcome such limitations \cite{martin2023sealclub}, and would still be applicable with the OCR features we define and present in this research. Another limitation is that, in the case of curved text relative to the x-y image processing grid, the `bounding box' for a word may not accurately reflect the word's center, which can provide some misalignment of keypoints found between images at different perspectives and lead to estimation error. 

\section{Concluding Remarks}

In this research, we presented a novel approach to document alignment using OCR outputs as features for homography estimation. By leveraging the spatial and textual information extracted from OCR, our method not only outperforms gradient-based descriptors designed for scene images, but also eliminates the need for original document images, making it suitable for scenarios with storage or privacy constraints. Further, the use of RANSAC ensures robustness to OCR noise, enabling accurate alignment even in the presence of recognition or matching errors. This approach has broad implications for document processing workflows, enabling efficient and scalable solutions for tasks like form processing, anomaly detection, and automated verification. 

By demonstrating the utility of document-domain models for geometric alignment, our method opens new possibilities for document analysis in real-world applications. Opportunities for future research include the iterative refinement of OCR matching following document alignment, including the joining of subword components which may be a match to a complete word in the registered template, and extension of spatial and textual OCR features to further vision tasks such as image unwarping or camera calibration by leveraging the expected geometric properties (e.g. kernings, linearity, etc.) of printed documents.


{
    \small
    \bibliographystyle{ieeenat_fullname}
    \bibliography{main}
}

\end{document}